\documentclass[letterpaper, 10 pt, conference]{ieeeconf}  % Comment this line out if you need a4paper

\IEEEoverridecommandlockouts                              % This command is only needed if 
\usepackage{algorithm}
\usepackage{algpseudocode}
\usepackage{amsmath}
\usepackage{amssymb}
\usepackage{xcolor} 
\usepackage{booktabs}
\usepackage{siunitx}
\usepackage{hyperref}
\usepackage{cleveref}
\usepackage{subcaption}
\usepackage{pifont}
\usepackage{colortbl}
\usepackage{threeparttable}
\usepackage{placeins}
\usepackage[table]{xcolor}
\usepackage{multirow}
\usepackage{graphicx}
\usepackage{cuted}
\usepackage{xspace}

\definecolor{chosengray}{RGB}{245,245,245}

\definecolor{bestperf}{HTML}{D5E8D4}
\definecolor{secondperf}{HTML}{C6DCEF}

\newcommand{\best}{\cellcolor{bestperf}\bfseries}
\newcommand{\second}{\cellcolor{secondperf}}

\title{\LARGE \bf
Don't Risk It!: Trajectory-Conditioned Distributional Risk for LiDAR-Only Crowd Navigation
}

\author{Mathai Mathew Pulicken$^{1\dagger}$, Laksh Nanwani$^{1}$, Basant Sharma$^{2}$, Phani Teja Singamaneni$^{3}$,\\ Arun Kumar Singh$^{2}$, K. Madhava Krishna$^{1}$% <-this % stops a space
\thanks{${\dagger}$ Corresponding Author.}
\thanks{$^{1}$ Robotics Research Center, IIIT Hyderabad, India.}%
\thanks{$^{2}$ University of Tartu, Estonia.}
\thanks{$^{3}$ Inria, Université de Lorraine, France.}
\thanks{Emails: (mathewp8616, lakshanshul, basantsharma1990, phaniteja.sp, aks1812)@gmail.com, mkrishna@iiit.ac.in}
}

\begin{document}

\maketitle

\begin{strip}
\vspace{-6.65em}

\centering
  \includegraphics[width=\linewidth]{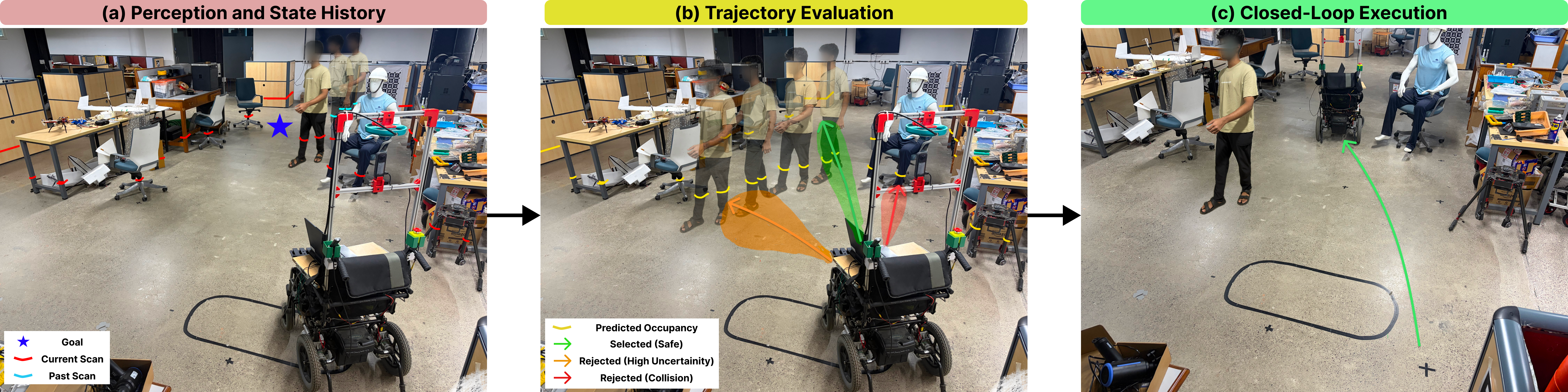}%

\captionof{figure}{\footnotesize{\textbf{Navigation via trajectory-conditioned distributional risk.} 
    \textbf{(a) Perception and State History} ($t = 0\,\text{s}$): Our framework operates strictly from raw 2D LiDAR grid history over the past $1.0\,\text{s}$ without object detection or pedestrian tracking, to forecast dynamic scene evolution. 
    \textbf{(b) Trajectory Evaluation:} Candidate rollouts are shaded by their predictive variance envelopes ($\pm \sigma$): maneuvers colliding with static obstacles are discarded (\textcolor{red}{red}); paths that cut aggressively across the pedestrian's anticipated path, triggering an elevated \textbf{risk penalty} (\textcolor{orange}{orange}); and the wide bypass corridor maintains tight predictive uncertainty, certifying safe passage (\textcolor{green!70!black}{green}). 
    \textbf{(c) Closed-Loop Execution:} The autonomous wheelchair executes the selected control sequence, avoiding the moving pedestrian.}}

\label{fig:teaser}
\end{strip}

\thispagestyle{empty}
\pagestyle{empty}

\newcommand{\papertitle}{DRIP}

%%%%%%%%%%%%%%%%%%%%%%%%%%%%%%%%%%%%%%%%%%%%%%%%%%%%%%%%%%%%%%%%%%%%%%%%%%%%%%%%

\begin{abstract}
Efficient crowd navigation requires anticipating dynamic collision-free space without relying on fragile object tracking. We propose the Don't Risk It Planner (\papertitle{}), an end-to-end LiDAR-only planning framework that couples occupancy forecasting with trajectory-conditioned clearance and uncertainty estimation. Operating directly on LiDAR-derived occupancy grid histories, \papertitle{} predicts a distribution over minimum robot–obstacle clearance for candidate control sequences to evaluate collision risk. The model is trained using a risk cost that grounds predictive uncertainty directly in empirical collision outcomes rather than geometric regression alone. In closed-loop experiments with up to 45 pedestrians, \papertitle{} achieves higher success rates with fewer collisions and reduced freezing behavior compared to state-of-the-art baselines.

\noindent
Project page: \href{https://dont-risk-it-planner.github.io}{https://dont-risk-it-planner.github.io}

\end{abstract}

%%%%%%%%%%%%%%%%%%%%%%%%%%%%%%%%%%%%%%%%%%%%%%%%%%%%%%%%%%%%%%%%%%%%%%%%%%%%%%%%
\section{INTRODUCTION}
\label{sec:introduction}

Safe, efficient crowd navigation requires a mobile robot to anticipate evolving space, as open paths can close and blocked routes may clear. Uncertain pedestrian dynamics and mutual occlusions~\cite{mun2023occlusion} complicate this reasoning. Crucially, robots must balance anticipatory spatial awareness with calibrated uncertainty estimates to prevent both collisions and over-conservative freezing~\cite{trevisan2025dynamic}.

Existing approaches separate this problem into pedestrian detection,
tracking, motion prediction, and planning. This decomposition introduces a multi-stage perception
dependency: detections must be associated over time before pedestrian
states can be estimated and their future motion predicted. Besides the
additional processing, incomplete observations can make these
intermediate estimates unreliable, allowing perception errors to
propagate into downstream decisions. 

An alternative approach is to forgo explicit pedestrian tracks and directly predict future occupancy maps. For example,~\cite{thomas2023foreseeable} presents a self-supervised learning-based approach for spatiotemporal occupancy prediction. More recently, SCOPE~\cite{xie2025scope} extends this direction through stochastic
occupancy forecasting and incorporates predicted occupancy and
uncertainty into navigation costmaps. These developments provide a basis
for anticipatory navigation without an object-centric forecasting
pipeline. Nevertheless, bypassing detection and tracking does not
eliminate uncertainty in the predicted scene. A central challenge remains:
how should uncertainty in occupancy forecasts be represented and learned
so that it informs the safety assessment of a candidate robot trajectory?

In this paper, we propose \papertitle{}, an end-to-end, LiDAR-only
planning framework that connects occupancy forecasting to
trajectory-conditioned clearance and uncertainty estimation. Our central
observation is that prediction errors matter differently depending on
the trajectory under consideration. An occupancy error far from the
robot's path may have little effect on its safety, whereas a small error
near its closest encounter with an obstacle may change whether the
trajectory should be selected. Consequently, improving scene-level
prediction accuracy alone need not improve the safety assessment most
relevant to planning. The broader principle of aligning prediction with
downstream risk has also been explored by~\cite{nishimura2023rap}, which learns
risk-biased motion forecasts. We pursue this alignment directly from
occupancy-level predictions by learning uncertainty in a
trajectory-specific safety quantity.

Specifically, as illustrated in Fig.~\ref{fig:teaser}, \papertitle{} forecasts scene evolution from a history of LiDAR-derived occupancy grids, without requiring explicit object detection or tracking. Given a candidate control sequence, a trajectory-conditioned model estimates a distribution over the minimum robot–obstacle clearance across the planning horizon. This representation expresses uncertainty in the closest encounter along the candidate trajectory, rather than only in individual occupancy predictions. The resulting clearance distribution is then mapped to collision risk, providing a common safety measure for comparing candidate motions.

We build upon \cite{sharma2025monompc} and train our forecasting model with a risk-calibrated objective that directly aligns the loss with downstream trajectory risk evaluation. By regularizing predicted uncertainty using supervision from safe and unsafe trajectories, our approach couples uncertainty estimation directly to planning safety rather than passive scene prediction. At inference, these risk estimates guide trajectory selection within a sampling-based local planner, balancing collision avoidance with goal progress.

We benchmark our approach in a closed-loop setting in dense environments with up to 45 pedestrians against TEB \cite{teb}, DWA \cite{DWA}, DRL-VO \cite{drlvo}, and SCOPE \cite{xie2025scope}. In extreme crowding (45 pedestrians), our framework achieves an $85.0\%$ success rate (a $31.8\%$ reduction in navigation failure over state-of-the-art predictive baselines), while reducing completion time by $14.5\%$ and improving velocity by $14.6\%$ compared to SCOPE. Under severe sensor noise, the performance gap widens substantially, cutting traversal time by over $50\%$ ($13.4\,\mathrm{s}$ vs. $27.0\,\mathrm{s}$) and maintaining an $86.0\%$ success rate.

\noindent In summary, the main contributions of this work are:
\begin{itemize}
    \item \textbf{A LiDAR-only predictive planning framework} that couples
    detection- and tracking-free occupancy forecasting with a
    trajectory-conditioned distribution over minimum clearance.

    \item \textbf{A risk-calibrated uncertainty-learning objective} that uses
    supervision from safe and unsafe trajectories to shape predictive variance
    through its downstream effect on collision-risk estimation.

    \item \textbf{Closed-loop validation in dense crowds and under sensor
    noise}, demonstrating fewer collisions and less freezing behavior,
    together with higher navigation success rates than state-of-the-art
    baselines.
\end{itemize}

\section{RELATED WORK}

\subsection{Crowd Navigation in Dynamic Environments}
Crowd navigation methods range from geometric collision avoidance, such as velocity obstacles~\cite{van2008reciprocal}, to learned pedestrian interaction models using structural RNNs, graphs, and attention mechanisms~\cite{singamaneni2024survey, chen2019crowd}. DRL-VO~\cite{drlvo} bridges these domains by combining velocity-obstacle rewards with LiDAR observations and pedestrian kinematics. Other methods incorporate risk by constructing local probabilistic risk maps~\cite{yang2023rmrl} or integrating barrier functions into learned policies~\cite{guo2025risk}. Recent works have also leveraged generative priors, such as VQ-VAEs~\cite{crowdsurfer} and conditional flow matching~\cite{crowdfm}, to model multimodal expert trajectories for sampling-based local planning. Within this broader effort, \papertitle{} predicts candidate motion safety directly from LiDAR histories. By learning clearance uncertainty for each candidate, it evaluates future encounters without requiring the explicit pedestrian-state estimates used by agent-based models and DRL-VO.

\subsection{Occupancy Forecasting for Predictive Navigation}
\noindent Occupancy maps provide a common representation for anticipating unseen
geometry~\cite{ramakrishnan2020occupancy}, learning social
costs~\cite{aisa2026general}, and predicting how occupied space evolves.
For dynamic navigation, LiDAR-based forecasting offers a way to
anticipate scene changes without detecting and tracking individual
pedestrians~\cite{thomas2023foreseeable}.
SOGMP and SOGMP++~\cite{xie2023stochastic} capture uncertainty by modeling
stochastic future occupancy. SCOPE ~\cite{xie2025scope}
further develops this approach by incorporating predicted occupancy and
uncertainty into navigation costmaps.
\papertitle{} builds on this tracking-free approach by connecting scene
forecast uncertainty to the safety of each candidate trajectory during training itself. In particular, it
learns a distribution over the closest robot--obstacle encounter across
the planning horizon. The next-step occupancy forecast supports immediate
collision checking, while the full forecast sequence informs this
trajectory-conditioned clearance model.

\subsection{Uncertainty-Aware and Distributional Planning}
Explicitly accounting for uncertainty allows robots to adjust speeds based on collision confidence~\cite{kahn2017uncertainty} and balance safety with goal progress under model uncertainty~\cite{nguyen2024uncertainty}. For crowd navigation, Ryu and Mehr~\cite{ryu2024integrating} manage distributional risk via CVaR approximations over pedestrian forecasts. Maximum Mean Discrepancy (MMD) provides another risk assessment tool, recently used to distill stochastic dynamics rollouts~\cite{sharma2025trajectory}. While this approach derives risk from stochastic dynamics, \papertitle{} computes squared MMD over a learned clearance distribution, conditioned on LiDAR histories and candidate controls, to measure how far safety-margin violations depart from an ideal zero-violation case.

\subsection{Learning Predictions for Downstream Risk}
\noindent Accurate predictions do not always lead to better planning decisions: an error near the robot's path may matter more than a larger error elsewhere. RAP~\cite{nishimura2023rap} addresses this mismatch by learning risk-biased forecasts of other agents' trajectories, so that expected costs approximate a desired risk measure.
MonoMPC~\cite{sharma2025monompc} provides a more direct foundation for our work by learning a control-conditioned minimum-clearance distribution for monocular navigation and supervising its uncertainty through an MMD-based risk model.
Building on MonoMPC, \papertitle{} extends clearance-based risk learning to dynamic crowds using temporal LiDAR observations and explicit future occupancy forecasting. Safe/unsafe trajectory labels supervise clearance uncertainty, while trajectory-ranking supervision encourages the occupancy decoder to preserve the relative safety of candidate paths. Together, these objectives connect both predictions to the safety decisions they support.

\begin{figure*}[t]
    \centering
    \includegraphics[width=.95\linewidth]{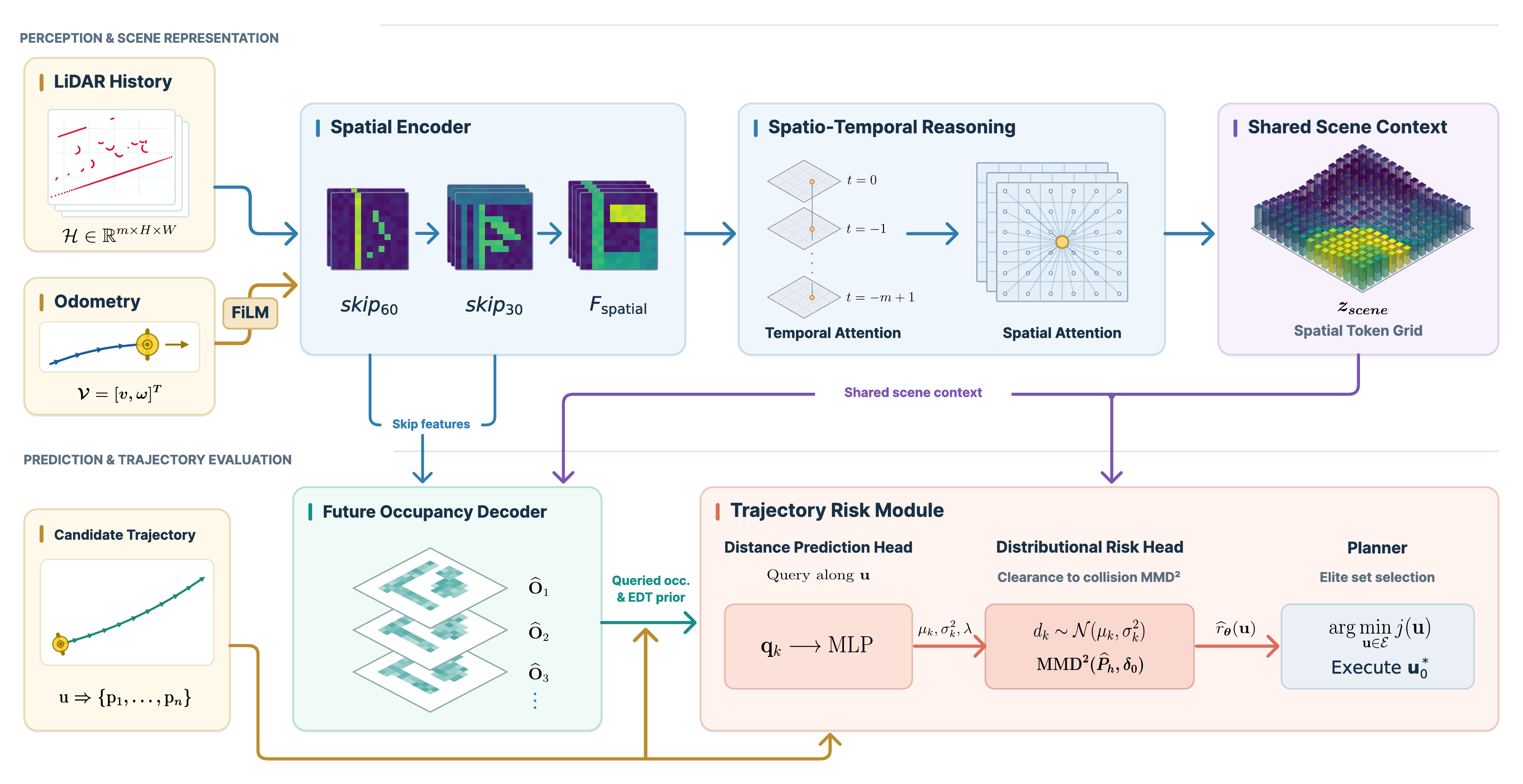}
    \caption{
    Overview of \papertitle{}'s LiDAR-only prediction and trajectory-conditioned
    risk architecture. A history $\mathcal{H}$ of pose-warped BEV occupancy maps is encoded independently by a shared-weight spatial CNN, conditioned on ego velocity $\mathcal{V}$ through FiLM. Spatiotemporal Transformer blocks then reason over the temporal history and spatial BEV tokens to produce a shared scene representation $z_{scene}$. Two branches use this representation: the occupancy decoder predicts the future evolution of the scene ($\widehat{\mathbf{O}}_k$) over a $20$-step horizon, while the trajectory-conditioned Distance Head combines scene features with candidate trajectory queries $\mathbf{q}_k$ to predict a stepwise clearance distribution $(\mu_k,\sigma_k^2)$ and a trajectory-specific kernel width $\lambda$. These predictions are subsequently used to compute the distributional MMD$^2$ risk $\widehat{r}_{\boldsymbol{\theta}}(\mathbf{u})$ for each candidate trajectory $\mathbf{u}$.
    }
    \label{fig:architecture}
    \vspace{-1mm}
\end{figure*}

\begingroup
  \setlength{\abovedisplayskip}{6pt plus 2pt minus 1pt}%
  \setlength{\belowdisplayskip}{6pt plus 2pt minus 1pt}%
  \setlength{\abovedisplayshortskip}{3pt plus 1pt minus 1pt}%
  \setlength{\belowdisplayshortskip}{3pt plus 1pt minus 1pt}%
  \setlength{\jot}{2pt}%

\section{Problem Formulation}
\label{sec:problem_formulation}

We consider a mobile robot navigating toward a goal among static
obstacles and moving pedestrians. At each planning cycle, a control
sequence $\mathbf{u}=[\mathbf{u}_0^\top,\ldots,\mathbf{u}_{n-1}^\top]^\top$
generates an $n$-step trajectory from the current state $\mathbf{x}_0$
through
\begin{equation}
    \mathbf{x}_{k+1}=\mathbf{f}(\mathbf{x}_k,\mathbf{u}_k),
    \qquad k=0,\ldots,n-1
    \label{eq:robot_dynamics}
\end{equation}
where $\mathbf{x}_k\in\mathbb{R}^{n_x}$ and
$\mathbf{u}_k\in\mathbb{R}^{n_u}$ are the state and input. Let
$\mathbf{p}_k=\mathbf{p}(\mathbf{x}_k(\mathbf{u}))\in\mathbb{R}^2$
denote the resulting planar positions and $\mathbf{g}\in\mathbb{R}^2$
the goal.

The prediction model receives LiDAR occupancy-grid history
$\mathcal{H}=(\mathbf{O}_{-m+1},\ldots,\mathbf{O}_0)$ and aligned robot
velocity history $\mathcal{V}=(\mathbf{v}_{-m+1},\ldots,\mathbf{v}_0)$,
where $\mathbf{v}_j=[v_j,\omega_j]^\top$ contains measured linear and
angular velocities. It forecasts occupancy maps
$\widehat{\mathbf{O}}_{1:n}$ in the current robot body frame. We exploit
the higher accuracy of the next-step forecast through the deterministic
collision cost
\begin{equation}
    j_{\mathrm{occ}}(\mathbf{u})
    =\ell_{\mathrm{occ}}\bigl(\mathbf{x}_1(\mathbf{u}),
                            \widehat{\mathbf{O}}_1\bigr)
    \label{eq:next_step_occupancy_cost}
\end{equation}
which penalizes footprint overlap with occupied space, holding
$\widehat{\mathbf{O}}_1$ fixed during optimization.

To assess safety over the full horizon, let $\mathcal{O}_k\subset\mathbb{R}^2$
denote the future obstacle-occupied region. For a circular robot of
radius $r_{\mathrm{rob}}$, define
\begin{align}
    d_k(\mathbf{u},\mathcal{O}_k)
      &=\inf_{\mathbf{q}\in\mathcal{O}_k}
        \|\mathbf{p}_k-\mathbf{q}\|_2-r_{\mathrm{rob}},
      \label{eq:instantaneous_clearance}\\
    d_{\min}(\mathbf{u})
      &=\min_{k=1,\ldots,n}d_k(\mathbf{u},\mathcal{O}_k)
      \label{eq:minimum_clearance}
\end{align}
Positive clearance indicates separation; nonpositive clearance indicates
contact or overlap. Uncertain future occupancy makes $d_{\min}$ random.
We learn its conditional law
$\mathcal{P}_{\boldsymbol{\theta}}(\cdot\mid\mathcal{H},\mathcal{V},\mathbf{u})$
and evaluate the risk
\begin{equation}
    \widehat{r}_{\boldsymbol{\theta}}(\mathbf{u})
      =\rho\!\left(\mathcal{P}_{\boldsymbol{\theta}}
        (\cdot\mid\mathcal{H},\mathcal{V},\mathbf{u})\right)
    \label{eq:risk_surrogate}
\end{equation}
where $\boldsymbol{\theta}$ denotes learned parameters and $\rho$ is a
distribution-based risk functional. We use the MMD construction in
Section~\ref{sec:methods} to model this.

For a candidate control sequence $\mathbf{u}$, the goal-reaching and smoothness costs are
\begin{equation}
    j_{\mathrm{goal}}=\|\mathbf{p}_n-\mathbf{g}\|_2^2,
    \quad
    j_{\mathrm{smooth}}=\sum_{k=0}^{n-1}
       \|\mathbf{u}_k-\mathbf{u}_{k-1}\|_2^2
    \label{eq:navigation_costs}
\end{equation}
where $\mathbf{u}_{-1}$ is the previously applied input. We solve
\begin{equation}
\begin{gathered}
  \begin{aligned}
    \mathbf{u}^{\star}
    =\underset{\mathbf{u}}{\operatorname{arg\,min}}\quad
    &w_g j_{\mathrm{goal}}(\mathbf{u})
      +w_s j_{\mathrm{smooth}}(\mathbf{u})\\
    &{}+w_o j_{\mathrm{occ}}(\mathbf{u})
      +w_r\widehat{r}_{\boldsymbol{\theta}}(\mathbf{u})
  \end{aligned}\\
  \begin{aligned}
    \text{s.t.}\quad
    &\mathbf{x}_{k+1}=\mathbf{f}(\mathbf{x}_k,\mathbf{u}_k),
      \quad\mathbf{x}_0=\mathbf{x}_{\mathrm{current}},\\
    &\mathbf{u}_k\in\mathcal{U},\quad
      \mathbf{x}_{k+1}\in\mathcal{X}.
  \end{aligned}
\end{gathered}
\label{eq:trajectory_optimization}
\end{equation}
The constraints hold for $k=0,\ldots,n-1$. The non-negative weights
balance the four costs, and $\mathcal{U},\mathcal{X}$ encode admissible
inputs and states. A sampling-based planner approximately solves this
problem, executes $\mathbf{u}_0^{\star}$, and replans with updated
observations.

\section{Methods}
\label{sec:methods}

Fig.~\ref{fig:architecture} summarizes our prediction pipeline. A
shared spatial encoder processes the LiDAR history $\mathcal{H}$,
with the aligned robot velocity history $\mathcal{V}$ conditioning its
feature channels through FiLM (feature-wise linear modulation) layers. Temporal attention combines features
across observations, and spatial attention captures relationships
across the bird's-eye-view (BEV) grid, yielding the shared scene representation
$\mathbf{z}_{scene}$.

Two connected branches use this representation. The occupancy decoder
combines $\mathbf{z}_{scene}$ with future-time embeddings and multi-scale skip
features to forecast occupancy maps $\widehat{\mathbf{O}}_{1:n}$. The trajectory-conditioned
branch queries scene features, occupancy forecasts, and their
Euclidean distance-transform priors along a candidate trajectory. These
queries combine scene context with local free-space geometry to predict stepwise clearance
means and variances, together with a trajectory-specific kernel
bandwidth. For each sampled clearance rollout, we take the minimum over
the horizon to obtain a sample of trajectory-level minimum clearance.
The risk module transforms these samples into safety-margin violations
and compares their empirical distribution with a point mass at zero.
It uses squared MMD with the predicted bandwidth $\lambda$ to produce the risk estimate
$\widehat{r}_{\boldsymbol{\theta}}(\mathbf{u})$. During training, we supervise $\widehat{\mathbf{O}}_{1:n}$ with ground truth occupancy. Moreover, following \cite{sharma2025monompc}, we compare the predicted risk $\widehat{r}_{\boldsymbol{\theta}}(\mathbf{u})$ with the ground-truth estimate and use the error to align the uncertainty prediction. 

We next present the building blocks of our pipeline, followed by the
training procedure and online planner.

\subsection{Spatiotemporal Scene Encoding and Occupancy Forecasting}
\label{subsec:scene_forecasting}

\subsubsection{Shared scene representation}
\label{subsubsec:scene_representation}
The observation history $\mathcal{H}$ contains $m$ LiDAR-derived
BEV occupancy grids. Past
scans are warped into the current robot body frame using measured
ego-motion. A spatial encoder with shared weights processes each grid
through convolutional stages at $60\times60$, $30\times30$, and
$15\times15$ resolutions. At the deepest stage, FiLM conditions the feature channels on the linear and
angular velocity history $\mathcal{V}$. This history provides the
robot-motion context aligned with the input LiDAR frames.

Temporal self-attention combines the $m$ observations independently at
each spatial cell. Spatial self-attention then processes the latest-time
feature grid to relate information across the scene. The resulting
scene representation $\mathbf{z}_{scene}$ encodes observed geometry and motion cues
without requiring pedestrian detections or persistent object identities.

\subsubsection{Occupancy forecasts for collision checking and clearance prediction}
\label{subsubsec:occupancy_decoder}
A convolutional decoder combines $\mathbf{z}_{scene}$, learned future-time
embeddings $\mathbf{e}_k$, and multi-scale features $\mathbf{F}_0$
from the latest observation to predict
\begin{equation}
    \widehat{\mathbf{O}}_k
    =f_{\mathrm{dec}}
       (\mathbf{z}_{scene},\mathbf{F}_0,\mathbf{e}_k),
    \qquad k=1,\ldots,n
    \label{eq:predicted_occupancy}
\end{equation}
where each $\widehat{\mathbf{O}}_k\in[0,1]^{h_{\mathrm{bev}}\times
w_{\mathrm{bev}}}$ is a soft-occupancy grid. Skip connections from $\mathbf{F}_0$ retain fine spatial detail during upsampling. We also
construct a Euclidean distance-transform grid $\mathrm{EDT}_k$ from each
forecast to provide a local free-space distance prior.

The first forecast $\widehat{\mathbf{O}}_1$ supplies the direct
next-step collision check in \eqref{eq:next_step_occupancy_cost}. For
each candidate, the check evaluates the robot footprint at its first
predicted state using this fixed map. The complete forecast sequence and
its distance-transform priors also provide geometric context for the
clearance model. Querying them along all future waypoints supports the
minimum-clearance distribution and full-horizon risk defined in
Section~\ref{sec:problem_formulation}.

\subsection{From Candidate Controls to Clearance and Risk}
\label{subsec:learning_clearance}

\subsubsection{Trajectory-conditioned clearance marginals}
\label{subsubsec:distance_head}
For a candidate $\mathbf{u}$, the dynamics in
\eqref{eq:robot_dynamics} determine the waypoint positions
$\mathbf{p}_k=\mathbf{p}(\mathbf{x}_k(\mathbf{u}))$,
$k=1,\ldots,n$. For querying, these positions are expressed in the
same %current body 
frame as the occupancy grids. At each waypoint, we
construct the feature vector,
\begin{equation}
\mathbf{q}_k=\bigl[
  \mathbf{z}_k^\top,\,
  \widehat{\mathbf{O}}_k(\mathbf{p}_k),\,
  \mathrm{EDT}_k(\mathbf{p}_k),\,
  \mathbf{f}_{\mathrm{skip}}(\mathbf{p}_k)^\top
\bigr]^\top
\label{eq:query_token}
\end{equation}
where $\mathbf{z}_k \in \mathbb{R}^{d}$ is the spatiotemporal latent token extracted at waypoint $\mathbf{p}_k$, $\widehat{\mathbf{O}}_k(\mathbf{p}_k)$ denotes the forecasted occupancy probability obtained via bilinear interpolation of the predicted $k$-th step BEV occupancy map at position $\mathbf{p}_k$, $\operatorname{EDT}_k(\mathbf{p}_k)$ is the Euclidean distance transform querying distance to static boundaries, and $f_{\mathrm{skip}}(\mathbf{p}_k)$ represents raw spatial coordinate skip-features.

A shared multilayer perceptron maps $\mathbf{q}_k$ to a mean $\mu_k$
and a positive variance $\sigma_k^2$, giving the conditional marginal
model
\begin{equation}
    d_k\mid\mathcal{H},\mathcal{V},\mathbf{u}
    \ \sim\ \mathcal{N}(\mu_k,\sigma_k^2),
    \qquad k=1,\ldots,n
    \label{eq:marginal_clearance}
\end{equation}
These are distributions of the footprint-adjusted clearance in
\eqref{eq:instantaneous_clearance}; the robot radius is therefore
already accounted for. We suppress the conditioning arguments of
$\mu_k$ and $\sigma_k$ for readability. A second head pools the query
features over the horizon to predict a positive kernel bandwidth
$\lambda_{\boldsymbol{\theta}}(\mathbf{u})$ for the risk computation
below. Its dependence on $\mathcal{H}$ and $\mathcal{V}$ is likewise
implicit.

\subsubsection{Minimum-clearance distribution}
\label{subsubsec:violation_distribution}
We obtain samples of trajectory-level minimum clearance by first
sampling the stepwise marginals and then taking the minimum within each
sampled rollout:
\begin{subequations}
\begin{alignat}{2}
    d_k^{(s)}&=\mu_k+\sigma_k\epsilon_k^{(s)},
    &\quad\epsilon_k^{(s)}&\overset{\mathrm{iid}}{\sim}\mathcal{N}(0,1)
    \label{eq:reparam_samples}\\
    d_{\min}^{(s)}(\mathbf{u})
    &=\min_{k=1,\ldots,n}d_k^{(s)},
    &s&=1,\ldots,n_s
    \label{eq:sample_min_clearance}
\end{alignat}
\end{subequations}

\subsubsection{Distributional risk from safety-margin violations}
\label{subsec:distributional_risk}
Let $d_{\mathrm{safe}}>0$ denote the required clearance buffer beyond
the robot footprint. For a candidate $\mathbf{u}$, the random variable
$h(\mathbf{u})=\max(0,d_{\mathrm{safe}}-d_{\min}(\mathbf{u}))$
measures how much this buffer is violated. The uncertainty in minimum
clearance therefore induces a distribution over safety-buffer
violations, which we denote by $P_h(\mathbf{u})$.

To evaluate this distribution, we transform the $n_s$ minimum-clearance
samples from \eqref{eq:sample_min_clearance} and form their empirical law:
\begin{align}
    h^{(s)}(\mathbf{u})
    &=\max\!\bigl(0,d_{\mathrm{safe}}-d_{\min}^{(s)}(\mathbf{u})\bigr)
    \label{eq:sample_violation}\\
    \widehat{P}_h(\mathbf{u})
    &=\frac{1}{n_s}\sum_{s=1}^{n_s}\delta_{h^{(s)}(\mathbf{u})}
    \label{eq:empirical_violation_distribution}
\end{align}
Each sample is zero when its rollout respects the buffer and positive
by the amount of the buffer shortfall otherwise. Together, these samples
form an empirical approximation of $P_h(\mathbf{u})$.
Here, $\delta_a$ denotes a unit point mass at $a$, so
$\widehat{P}_h$ assigns equal weight $1/n_s$ to each sampled violation.
The hat identifies this sample-based representation. An ideal safe
trajectory has zero violation for every realization, giving the target
distribution $\delta_0$.

We measure discrepancy from this target using squared MMD with a Laplacian kernel,
$k_\lambda(a,b)=\exp(-|a-b|/\lambda)$. For fixed observations and
candidate $\mathbf{u}$, write
$\lambda=\lambda_{\boldsymbol{\theta}}(\mathbf{u})$. Since the target is a point mass, the empirical risk has the explicit form
\begin{align}
    \widehat{r}_{\boldsymbol{\theta}}(\mathbf{u})
    &=\operatorname{MMD}_{k_\lambda}^{2}
      \bigl(\widehat{P}_h(\mathbf{u}),\delta_0\bigr)
    \label{eq:mmd_definition}\\
    &=\frac{1}{n_s^2}\sum_{s,t=1}^{n_s}
      \exp\!\bigl(-|h^{(s)}-h^{(t)}|/\lambda\bigr)\notag\\
    &\quad-\frac{2}{n_s}\sum_{s=1}^{n_s}
      \exp\!\bigl(-h^{(s)}/\lambda\bigr)+1
    \label{eq:mmd_closed_form}
\end{align}
Equation~\eqref{eq:mmd_closed_form} evaluates the empirical MMD directly
from the violation samples. In the notation of
\eqref{eq:risk_surrogate}, $\rho$ includes the transformation from
clearance to violation and uses the candidate's predicted bandwidth.

\subsection{Risk-Calibrated Training}
\label{subsec:training_and_losses}
\noindent  We build upon \cite{sharma2025monompc} to develop a training pipeline that regularizes uncertainty not only through the forecasting error but also through how it affects the downstream planning risk. This is done in two ways: (i) by aligning predicted risk during training with the ground-truth risk and (ii) by a ranking loss that differentiates between safe and unsafe trajectory proposals.

\subsubsection{Stage 1: occupancy forecasting}
\label{subsubsec:stage1_training}
We first optimize the spatial encoder, attention modules, and occupancy
decoder using ground-truth future occupancy grids
$\mathbf{O}_{1:n}^{\mathrm{gt}}$:
\begin{equation}
\mathcal{L}_{\mathrm{recon}}
 =\bigl(\mathcal{L}_{\mathrm{focal}}+\mathcal{L}_{\mathrm{Dice}}\bigr)
   (\widehat{\mathbf{O}}_{1:n},\mathbf{O}_{1:n}^{\mathrm{gt}})
\label{eq:recon_loss}
\end{equation}
Focal and Dice loss functions address the imbalance between occupied and free
cells. To emphasize changes in occupancy, reconstruction supervision
uses spatial weights
\begin{equation}
    \mathbf{W}_k(\mathbf{p})=1+\beta_{\mathrm{dyn}}\,
    \mathbb{I}\!\left[
      \mathbf{O}_k^{\mathrm{gt}}(\mathbf{p})\ne\mathbf{G}_0(\mathbf{p})
    \right]
    \label{eq:dynamic_weighting}
\end{equation}
with $\beta_{\mathrm{dyn}}=4$. The current and future grids are
expressed in the same frame. This weighting emphasizes cells that become
occupied or free, reducing the dominance of unchanged background
geometry in the training signal.

\subsubsection{Stage 2: clearance and risk supervision}
\label{subsubsec:stage2_training}
We freeze the spatial encoder and attention modules, retain
reconstruction supervision for the occupancy decoder, and optimize
\begin{equation}
\begin{aligned}
    \mathcal{L}_{\mathrm{stage2}}
    &=\mathcal{L}_{\mathrm{recon}}
      +\beta_{\mathrm{dist}}\mathcal{L}_{\mathrm{dist}}\\
    &\quad+\beta_{\mathrm{risk}}\mathcal{L}_{\mathrm{risk}}
      +\beta_{\mathrm{haz}}\mathcal{L}_{\mathrm{haz}}
\end{aligned}
\label{eq:stage2_total_loss}
\end{equation}
where the non-negative $\beta$ coefficients weight the objectives.

\noindent\textit{Clearance likelihood:}
For each sampled candidate, ground-truth stepwise clearances
$d_k^{\mathrm{gt}}(\mathbf{u})$ are computed using the same robot
footprint convention as \eqref{eq:instantaneous_clearance}. Omitting an
additive constant, the Gaussian negative log-likelihood is
\begin{equation}
\mathcal{L}_{\mathrm{dist}}
 =\frac{1}{2n}\sum_{k=1}^{n}
   \bigl[(d_k^{\mathrm{gt}}-\mu_k)^2/\sigma_k^2+\log\sigma_k^2\bigr]
\label{eq:gaussian_nll}
\end{equation}
The residual term penalizes inaccurate means relative to the predicted
variance, while the logarithmic term discourages explaining every error
by increasing the variance.

\noindent\textit{Risk-based trajectory supervision:}
The ground-truth minimum clearance defines the binary label
\begin{equation}
y(\mathbf{u})=\mathbb{I}\!\bigl[
  \min\nolimits_{1\leq k\leq n}d_k^{\mathrm{gt}}(\mathbf{u})<d_{\mathrm{safe}}
\bigr]
\label{eq:unsafe_label}
\end{equation}
where $y=1$ denotes a safety-buffer violation. A lightweight classifier
$g_{\boldsymbol{\phi}}$ takes the empirical MMD score and violation
statistics as input:
\begin{subequations}
\begin{align}
    \ell(\mathbf{u})
    &=g_{\boldsymbol{\phi}}\!\left(
      \widehat{\mathbf{r}}_{\boldsymbol{\theta}}(\mathbf{u}),
      \overline{h},\sigma_h,h_{\max}\right),
      \label{eq:risk_logit}\\
    \mathcal{L}_{\mathrm{risk}}
    &=\operatorname{BCEWithLogits}\bigl(\ell(\mathbf{u}),y(\mathbf{u})\bigr)
      \label{eq:risk_ce_loss}
\end{align}
\end{subequations}
Here, $\overline{h}$ is the average constraint violation residual,
$\sigma_h$ is the standard deviation of these residuals, and
$h_{\max}$ is the worst-case constraint violation across the sampled
rollouts. These statistics summarize the average severity, variability,
and largest violation associated with the candidate trajectory.
The classifier parameters $\boldsymbol{\phi}$ are used only during
training. Gradients pass through the violation statistics and MMD
computation to train the clearance head and bandwidth predictor.
Consequently, the uncertainty receives supervision through its effect
on distinguishing safe and unsafe trajectories, in addition to the
likelihood objective.

\paragraph{Auxiliary Hazard Supervision}
To condition the future occupancy decoder beyond passive pixel reconstruction, we introduce an auxiliary hazard-margin loss evaluated over candidate path queries. For each trajectory $\mathbf{u}$, we bilinearly interpolate the soft occupancy probability at each waypoint and compute the scalar hazard cost $c(\mathbf{u})$ by averaging the highest values over a $0.5\,\text{s}$ window. For trajectory pairs where path $i$ is strictly safer than path $j$ in ground truth ($c_{\mathrm{gt}}(\mathbf{u}^{(i)}) < c_{\mathrm{gt}}(\mathbf{u}^{(j)})$), we penalize the decoder if the predicted hazard separation falls below margin $m_{\mathrm{haz}} = 0.05$:
\begin{equation}
    \mathcal{L}_{\mathrm{haz}} = \frac{1}{|\mathcal{P}|} \sum_{(i,j) \in \mathcal{P}} \max\big(0, \, c_{\mathrm{pred}}(\mathbf{u}^{(i)}) - c_{\mathrm{pred}}(\mathbf{u}^{(j)}) + m_{\mathrm{haz}}\big).
    \label{eq:hazard_margin_loss}
\end{equation}
This directly conditions the occupancy decoder, focusing gradient updates on navigable path boundaries rather than passive background cells.

\noindent\textit{Gradient routing:}
During Stage 2, scene features, skip features, occupancy values, and
distance-transform priors are detached before they enter the trajectory
querying branch. The likelihood loss $\mathcal{L}_{\mathrm{dist}}$
trains the trajectory-query and clearance modules.
The risk loss $\mathcal{L}_{\mathrm{risk}}$ also updates these modules
and additionally trains the bandwidth predictor and classifier
$g_{\boldsymbol{\phi}}$. Neither loss updates perception through this
branch. The occupancy decoder is updated by
$\mathcal{L}_{\mathrm{recon}}$ and $\mathcal{L}_{\mathrm{haz}}$
through its undetached outputs. The encoder and attention modules remain
frozen. This routing preserves the shared scene representation while allowing both downstream objectives to refine their respective branches.

\subsection{Sampling-Based Optimization and Learned Model Inference}
\label{subsec:online_planning}
\noindent We solve \eqref{eq:trajectory_optimization} using DWA-style trajectory
sampling. At each cycle, the histories $(\mathcal{H},\mathcal{V})$
produce the shared representation $\mathbf{z}_{scene}$, occupancy forecasts
$\widehat{\mathbf{O}}_{1:n}$, and distance-transform priors once.
We sample $n_c$ admissible control sequences and roll them out from
$\mathbf{x}_0$ subject to the dynamics and state limits.
Using the cached perception outputs, we evaluate next-step occupancy
costs and trajectory-conditioned risk in a batch, with shared
clearance-noise samples and fixed learned parameters. We adopt a two-stage approach. We first select a subset of sampled trajectories with the lowest safety cost (one-step occupancy+risk, recall \eqref{eq:navigation_costs}). We then evaluate the full cost over this subset to choose the lowest-cost trajectory for execution.

\endgroup

\section{Validation and Benchmarking}
\label{sec:experiments}

We evaluate \papertitle{} against state-of-the-art classical, optimization-based, and learning-based navigation planners in simulated indoor environments with dynamic crowds. Our experiments investigate two primary questions:
\begin{enumerate}[leftmargin=*]
    \item \textbf{Scalability Across Crowd Densities:} Does trajectory-conditioned distributional risk improve navigation success and robot behavior in high pedestrian density cases?
    \item \textbf{Robustness to Sensor Degradation:} Does evaluating continuous clearance distributions maintain performance under LiDAR range noise where classical costmap planners degrade?
\end{enumerate}

\subsection{Implementation and Experimental Setup}
\label{subsec:experimental_setup}

\paragraph{Robot Platform and Simulation Environment}
We use a differential-drive TurtleBot2 mobile base with a forward-facing 2D LiDAR modeled after the Hokuyo UTM-30LX, with a $270^\circ$ field of view. We conduct simulated evaluations in the PEDSIM Simulator \cite{gloor_pedsim_2016}, evaluating the Lobby world with 35 and 45 active pedestrians.

\paragraph{Data Collection and Trajectory Supervision}
To train the model, we construct a hybrid dataset comprising:
\begin{itemize}[leftmargin=*, noitemsep, topsep=0pt]
    \item \textbf{Synthetic Simulation ($\sim\mathbf{500\mathbf{k}}$):} Room, corridor, and intersection topologies simulated via PySocialForce~\cite{pysocialforce}.
    \item \textbf{Real-World Data ($\sim\mathbf{750\mathbf{k}}$):} Extracted from $\approx 75$ hours of HuRoN~\cite{hirose2023sacson} and $\approx 5$ hours of SCAND~\cite{karnan2022socially} logs to capture authentic human motion and sensor noise.
\end{itemize}
To supervise our trajectory-conditioned risk and ranking objectives, raw LiDAR point clouds from these datasets are ego-motion compensated and projected into consecutive BEV grids registered to the latest robot pose.

\paragraph{Baselines and Evaluation Protocol}
We benchmark \papertitle{} against four leading navigation policies under identical dynamic limits. Crucially, while \papertitle{}, DWA, TEB, and SCOPE operate strictly from raw 2D LiDAR observations without explicit agent tracking, DRL-VO assumes privileged dynamic state inputs.
Performance is measured using four standard criteria: (i) \textbf{Success Rate (\%)} (reaching the goal without collision or timeout, $T_{\max} = 90\,\text{s}$); (ii) \textbf{Average Time (s)} to completion; (iii) \textbf{Average Path Length (m)}; and (iv) \textbf{Average Speed (m/s)}.

\begin{table*}[!t]
\centering

% \scriptsize
\footnotesize
%\small
\setlength{\tabcolsep}{4.5pt}
\renewcommand{\arraystretch}{1.2}
\vspace{1.6mm}

% --- Eliminates booktabs gaps so horizontal and vertical lines intersect ---
\setlength{\aboverulesep}{0pt}
\setlength{\belowrulesep}{0pt}

\begin{tabular}{llcccc|cccc}
\toprule
& & \multicolumn{4}{c|}{\textbf{Nominal Navigation (100 runs each)}} & \multicolumn{4}{c}{\textbf{Sensor Noise Degradation ($\sigma = 0.05\,\mathrm{m}$) (50 runs each)}} \\
\cline{3-10}
\textbf{Density} & \textbf{Planner}
& \textbf{Succ. (\%) $\uparrow$} & \textbf{Time (s) $\downarrow$} & \textbf{Path (m) $\downarrow$} & \textbf{Vel. (m/s) $\uparrow$}
& \textbf{Succ. (\%) $\uparrow$} & \textbf{Time (s) $\downarrow$} & \textbf{Path (m) $\downarrow$} & \textbf{Vel. (m/s) $\uparrow$} \\
\midrule

\multirow{6}{*}{\textbf{35 ped}}
& DRLVO$^\dagger$   & 74.0          & 11.5 & 5.39 & 0.460 & --            & --   & --   & --    \\
& DWA               & 79.0          & 13.5 & 5.08 & 0.355 & 38.0          & 31.8 & 7.33 & 0.195 \\
& SCOPE             & \second{80.0} & 14.8 & 5.42 & 0.365 & \second{66.0}          & 26.3 & 6.10 & 0.227 \\
& TEB               & 71.0          & 12.3 & 5.46 & 0.413 & 30.0          & 40.2 & 18.09& 0.368 \\
& \textbf{Ours} & \best{87.0}   & 12.1 & 4.83 & 0.394 & \best{82.0}   & 12.9 & 4.67 & 0.357 \\
\midrule

\multirow{6}{*}{\textbf{45 ped}}
& DRLVO$^\dagger$   & \second{82.0} & 11.5 & 5.42 & 0.455 & --            & --   & --   & --    \\
& DWA               & 74.0          & 14.7 & 5.47 & 0.352 & 52.0          & 28.8 & 5.60 & 0.175 \\
& SCOPE             & 78.0          & 15.2 & 5.09 & 0.328 & \second{64.0}          & 27.0 & 5.53 & 0.204 \\
& TEB               & 76.0          & 12.5 & 5.38 & 0.413 & 20.0          & 46.5 & 21.01& 0.312 \\
& \textbf{Ours} & \best{85.0}   & 13.0 & 4.82 & 0.376 & \best{86.0}   & 13.4 & 4.87 & 0.356 \\

\bottomrule
\end{tabular}
\caption{Closed-loop navigation performance and sensor noise robustness in the simulated Lobby world with 35 and 45 pedestrians. Higher success rate ($\uparrow$) and velocity ($\uparrow$) are better; lower completion time ($\downarrow$) and path length ($\downarrow$) are better. \colorbox[HTML]{D5E8D4}{\textbf{Best}} and \colorbox[HTML]{C6DCEF}{\textbf{second-best}} success rates are highlighted. ($^\dagger$DRL-VO utilizes privileged pedestrian tracking/velocities and is excluded from direct sensor noise testing).}
\label{tab:unified_benchmarks}

\vspace{-1mm}
\end{table*}

\subsection{Validation of the Learned Risk Model}
\begin{figure}[htbp]
    \centering
        \includegraphics[width=0.75\linewidth]{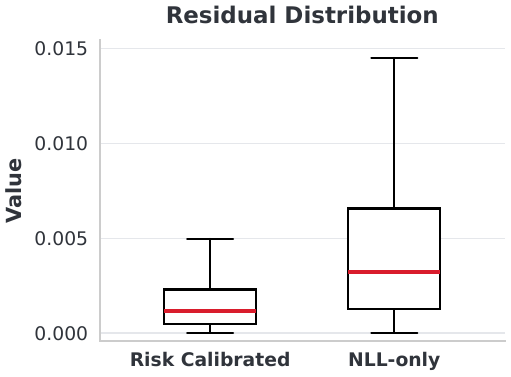}
        \caption{Box plot of best-of-$K$ clearance residuals relative to privileged ground truth across dynamic evaluation episodes. The risk-calibrated model (\eqref{eq:stage2_total_loss}) yields a lower median error and narrower spread than the NLL-only baseline without $\mathcal{L}_{risk}$, $\mathcal{L}_{haz}$), demonstrating well-calibrated predictive variance that tightly bounds actual obstacle encounters.}
        \label{fig:residual_boxplot}
 % \vspace{-2mm}
\end{figure}

To validate our risk-calibrated training objective, we measure the best-of-$K$ residual between sampled trajectory clearances and the ground-truth minimum clearance. As illustrated in Fig.~\ref{fig:residual_boxplot}, our risk-calibrated model produces a significantly lower median error and tighter inter-quartile range than the NLL-only baseline. Grounding uncertainty estimation in downstream collision risk directly improves the accuracy of the learned distances, ensuring both the predicted clearance boundaries closely match actual pedestrian encounters, and reliable navigation even in dense pedestrian traffic.

\subsection{Simulation Benchmarks: Density Scalability and Noise Robustness}
\label{subsec:sim_benchmarks}

Table~\ref{tab:unified_benchmarks} evaluates closed-loop navigation performance and sensor noise resilience in the simulated Lobby world under high crowd densities (35 and 45 active pedestrians).

\paragraph{Nominal Navigation and Crowd Scalability}
Under dense crowd interaction, classical geometric planners like DWA and TEB degrade substantially. SCOPE partially mitigates collisions via predicted occupancy, but translating stochastic forecasts into static Gaussian inflation layers induces excessive conservatism, leading to the slowest completion times. Although the privileged DRL-VO achieves high forward speeds using ground-truth agent tracking, its lack of explicit metric clearance constraints leads to collisions. In contrast, \papertitle{} achieves the highest overall success rate because the deterministic Euclidean distance transform penalty prevents collisions against static walls while the learned MMD risk navigates dynamic human flow.

\paragraph{Resilience Under Sensor Degradation}
Injecting independent Gaussian noise ($\sigma = 0.05\,\mathrm{m}$) into raw LiDAR ranges exposes fundamental failure modes in costmap-based navigation. DWA and TEB have a significant drop in performance due to added noise, as the noise causes DWA to repeatedly trigger emergency braking and destabilizes TEB's elastic-band optimization. SCOPE filters some point noise, but range uncertainty compounds costmap inflation. Conversely, \papertitle{} maintains robust performance under identical corruption, almost matching its noise-free baseline, while preserving direct trajectories. This noise immunity is due to our trajectory risk, as our distance head predicts higher variance $\sigma_k^2$ for these noisy point clouds, which then increases MMD risk $\widehat{r}_{\boldsymbol{\theta}}(\mathbf{u})$ via \eqref{eq:mmd_closed_form}. Rather than emergency braking, \papertitle{} hedges against observation uncertainty by committing to less crowded paths while maintaining continuous forward progress.

\subsection{Qualitative Forecasting Analysis}
\begin{figure}[htbp]
    \centering
    \includegraphics[width=\linewidth]{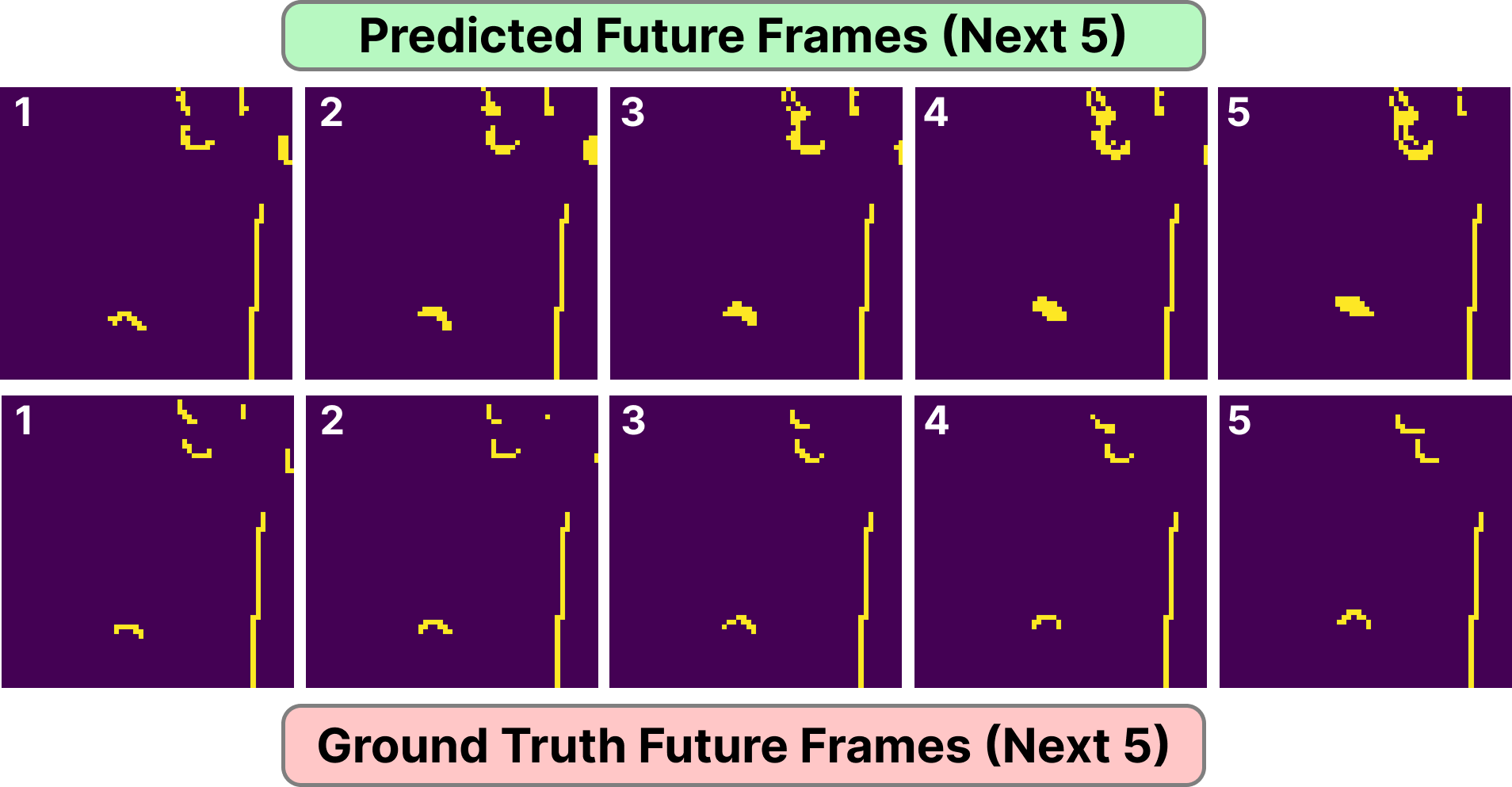}
    \caption{Visualization of the occupancy forecasting. \textit{Top row:} The first 5 steps of the predicted future occupancy maps ($\widehat{\mathbf{O}}_k$), visualized as thresholded binary maps, taken from the full 20-step planning horizon. \textit{Bottom row:} The corresponding ground-truth future occupancy maps ($\mathbf{O}_k^{\mathrm{gt}}$). The network successfully anticipates the continuous motion of dynamic obstacles and maintains static boundary structures, closely matching the ground truth without relying on explicit pedestrian tracking.}
    \label{fig:qualitative_forecasting}
% \vspace{-2mm}
\end{figure}

To understand the planner's high success rate, we qualitatively evaluate the underlying spatiotemporal forecasting module. Fig.~\ref{fig:qualitative_forecasting} illustrates the model's ability to anticipate scene evolution strictly from LiDAR histories. Despite lacking explicit pedestrian tracking or data association, the network accurately forecasts the continuous motion of dynamic obstacles while preserving static boundary structures. This tracking-free prediction directly enables the distributional risk assessment used by the local planner.

\subsection{Runtime Analysis}
\label{subsec:runtime_analysis}

To verify real-time feasibility for closed-loop deployment, we benchmark the complete inference and planning pipeline on a mobile workstation equipped with an NVIDIA GeForce RTX 3060 Laptop GPU. The local planner targets a strict $10\,\mathrm{Hz}$ control loop while evaluating $N_{\mathrm{cand}}$ kinematically unrolled trajectory candidates over a $2.0\,\mathrm{s}$ planning horizon. The entire decision loop executes in about $70\,\mathrm{ms}$, which is well within the $100\,\mathrm{ms}$ cycle threshold. This computational headroom ensures consistent real-time re-planning while leaving ample processing budget for onboard state estimation and low-level motor actuation.

\subsection{Real-World Hardware Deployment}
\label{subsec:real_world_experiments}

To demonstrate real-world applicability and sim-to-real transfer, we deploy our navigation framework on two distinct robotic platforms with differing form factors and sensing modalities. For both platforms, the complete perception and planning pipeline was executed entirely onboard using the mobile workstation described in Section~\ref{subsec:runtime_analysis}:
\begin{enumerate}[leftmargin=*]
    \item \textbf{Pioneer 3-DX Mobile Robot:} A compact differential-drive research platform equipped with a 2D RPLIDAR A3 scanner.
    \item \textbf{Autonomous Wheelchair:} A custom full-scale smart wheelchair platform (pictured in Fig.~\ref{fig:teaser}) equipped with a Livox Mid-360 3D LiDAR. 3D point clouds are filtered, ground-segmented, and projected into dynamic 2D BEV occupancy grids registered to the latest robot pose.
\end{enumerate}

\paragraph{Deployment Scenarios and Results}
We conducted field evaluations across two demanding indoor environments: a cluttered laboratory space with narrow passages, and an academic corridor characterized by bi-directional pedestrian flows and unpredictable group motion. Across 20 total closed-loop trials (spanning path lengths of 15 to 35 m) where humans acted as unscripted dynamic obstacles, \papertitle{} achieved an $85.0\%$ success rate (17 out of 20 runs) without requiring human intervention. In open corridors, the planner proactively committed to wider bypass paths around walking pedestrians well before mutual deceleration thresholds. In tight hallway bottlenecks, the decoupled clearance safety floor prevented the base from grazing walls while the learned MMD risk safely moderated forward speed through moving crowd pockets, avoiding freezing deadlock.

\section{Conclusion and Future Work}
\label{sec:conclusion}

In this work, we showed that conventional crowd navigation paradigms exhibit fundamental failure modes in dense environments: classical geometric costmaps inflate dynamic obstacles into impassable barriers, while end-to-end learning methods either lack formal safety margins or depend on privileged multi-agent tracking pipelines that fail in crowded scenes. We addressed these limitations by introducing \papertitle{}, a framework that formulates safe crowd navigation as a trajectory-conditioned distributional risk evaluation problem directly over raw LiDAR observations. 

Closed-loop benchmarks in simulation demonstrate that \papertitle{} maintains high success rates in crowded scenes, outperforming both classical planners (DWA, TEB) and privileged tracking-based reinforcement learning (DRL-VO). Furthermore, evaluation under synthetic sensor degradation reveals that when exposed to LiDAR coordinate jitter, \papertitle{} preserves safe paths without much performance drop, whereas other methods are much more brittle. Finally, real-world deployments on multiple robots confirm safe planning across various real-world scenes.

A primary limitation of our current framework is its reliance on a discrete set of unrolled trajectory primitives, which can restrict maneuver expressiveness in complex, topologically confined geometries. Future work will investigate integrating continuous generative trajectory priors directly with our distributional MMD risk metric to combine expressive path generation with rigorous safety certification.

\section*{ACKNOWLEDGMENT}
While LLMs (ChatGPT, Gemini, Claude) were used for text refinement and scripting, all intellectual contributions, formulations, and evaluations are original.


%%%%%%%%%%%%%%%%%%%%%%%%%%%%%%%%%%%%%%%%%%%%%%%%%%%%%%%%%%%%%%%%%%%%%%%%%%%%%%%%

%%%%%%%%%%%%%%%%%%%%%%%%%%%%%%%%%%%%%%%%%%%%%%%%%%%%%%%%%%%%%%%%%%%%%%%%%%%%%%%%

%%%%%%%%%%%%%%%%%%%%%%%%%%%%%%%%%%%%%%%%%%%%%%%%%%%%%%%%%%%%%%%%%%%%%%%%%%%%%%%%

\bibliographystyle{IEEEtran}
\bibliography{references}

\end{document}